\documentclass[10pt, a4paper]{article}
\usepackage{hybridreport}
\usepackage[square,authoryear]{natbib}  % author-year citations, as corl2026_yubi

\usepackage{multirow}
\usepackage{makecell}
\usepackage{tikz}
\usepackage{cleveref}               % \Cref -- must be loaded AFTER hyperref
\crefname{section}{Sec.}{Secs.}
\Crefname{section}{Sec.}{Secs.}
\crefname{subsection}{Sec.}{Secs.}
\Crefname{subsection}{Sec.}{Secs.}
\crefname{subsubsection}{Sec.}{Secs.}
\Crefname{subsubsection}{Sec.}{Secs.}
\crefname{figure}{Fig.}{Figs.}
\Crefname{figure}{Fig.}{Figs.}
\usepackage{xspace}

\DeclareTextFontCommand{\textbf}{\bfseries}

\graphicspath{{figs/}}
\newcommand{\grouplabel}[1]{\textcolor{black!55}{\textit{#1}}}  % method-group headers in tables

\newcommand{\dashedrow}[1]{%
  \multicolumn{#1}{@{}c@{}}{\leavevmode\leaders\hbox to 2pt{\hss\vrule height \dimexpr0.8ex+0.2pt\relax depth -\dimexpr0.8ex-0.2pt\relax width 1pt\hss}\hskip 0pt plus 1filll\kern0pt}%
  \\[-1.3ex]}

\newcommand{\eg}{e.g.,\xspace}

\newcommand{\revised}[1]{#1}

\def\DBhours{8434\xspace}      % 8434:08:36 of recorded interaction
\reporttype{Technical Report}
\logo{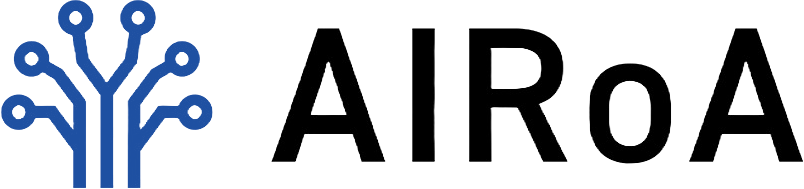}               % symlink to template/logo.png, shown in the first-page header
\title{YUBI-STAG: Contact and Semantic-Rich Alignment for VLAs via Automated Video-Language Grounding}
\shorttitle{YUBI-STAG: Contact and Semantic-Rich Alignment for VLAs}

\author{
  \mbox{Masatoshi Tateno\textsuperscript{1,2\textdagger}}\quad
  \mbox{Takehiko Ohkawa\textsuperscript{1\textdaggerdbl}}\quad
  \mbox{Yueh-Hua Wu\textsuperscript{1}}\quad
  \mbox{Hanlong Li\textsuperscript{1,3}}\quad
  \mbox{Tatsuya Matsushima\textsuperscript{1,2}}\quad
  \mbox{Yoichi Sato\textsuperscript{2}}\quad
  \mbox{Kei Ota\textsuperscript{1}}
}

\institution{
  \textsuperscript{1}AI Robot Association (AIRoA)\quad
  \textsuperscript{2}The University of Tokyo\quad
  \textsuperscript{3}Institute of Science Tokyo
}

\date{October 2026}
\correspondence{%
  \href{mailto:tateno.masatoshi@airoa.org}{\texttt{tateno.masatoshi@airoa.org}}\\
  Project page: \url{https://research.airoa.io/publications/yubi-stag/}\\
  \textsuperscript{\textdagger}VLM Lead\quad
  \textsuperscript{\textdaggerdbl}VLA Lead%
}

\abstracttext{

  Vision-Language-Action (VLA) models acquire broad manipulation capabilities via large-scale pretraining, yet eliciting them through language requires fine-grained alignment between instructions and physical interactions. Existing robot demonstrations, however, typically provide only coarse task descriptions, omitting how actions are executed, including which gripper acts, which object is contacted, and how it is grasped and moved.
We introduce \emph{YUBI-STAG}, a framework for \revised{\textbf{S}patio-\textbf{T}emporal \textbf{A}nnotation and \textbf{G}rounding} that automatically enriches manipulation demonstrations with interaction-rich semantics for aligning pretrained VLAs with fine-grained manipulation language. Combining contact-object segmentation with vision-language models, YUBI-STAG annotates object identities, attributes and states, per-gripper actions, bimanual coordination, and spatially grounded interactions. \revised{YUBI-STAG assumes a temporally localized action sequence and relies on multi-stage VLM inference. To remove these requirements, we distill the pipeline into \emph{YUBI-VLM}, which directly recovers the action structure and corresponding annotations from raw, unsegmented video in few inference calls.} It also operates from wrist views alone, enabling portable annotation without fixed external cameras. \revised{We evaluate both YUBI-STAG and YUBI-VLM on our new \emph{YUBI-STAG-Bench}, which measures annotation quality across temporal, semantic, and spatial grounding tasks. Results show that the distilled YUBI-VLM retains much of YUBI-STAG's annotation accuracy with fewer inference calls and shorter runtime while generalizing to unseen manipulations.}
Finally, we use the resulting annotations to post-train YUBI-pretrained VLA policies, aligning their manipulation capabilities with fine-grained language and contact-aware interaction structure. Experiments on bimanual manipulation show improved manipulation performance and instruction following, including control over object identity, acting gripper, target location, and spatial relations absent from the original demonstration labels. \revised{Our results show that richly grounded interaction annotations can better elicit pretrained VLA capabilities through language.}
}

\begin{document}

\maketitle

\section{Introduction}

A robot demonstration shows how the task was executed, but its label records
only what the task was.
A session of sorting industrial parts into bins shows which gripper acts, which
part it contacts, and how that part is grasped, handed between grippers, and
released.
However, all of it is annotated as ``pick a part and place it in a box'', which omits interaction details.
This matters because vision-language-action (VLA) models acquire broad
manipulation capability from large-scale
pretraining~\citep{brohan2023rt2,kim2024openvla,octo2024,black2024pi0,black2025pi05},
and language is the only channel through which that capability is elicited.
A language label therefore does more than document an episode, while current
datasets include one task-level label per episode and repeat it across every
demonstration of the same task class.
For instance, recent robot-free handheld interfaces such as
UMI~\citep{chi2024universal} and YUBI~\citep{ohkawa2026yubi} offer large-scale data
(\eg \DBhours~hours of bimanual interaction by YUBI) at higher throughput than
teleoperation systems~\citep{zhao2023learning,fu2024mobilealoha}.
Yet fine granularity and diversity of that language remain the bottleneck.

\begin{figure}[t]
  \centering
  \includegraphics[width=0.8\columnwidth]{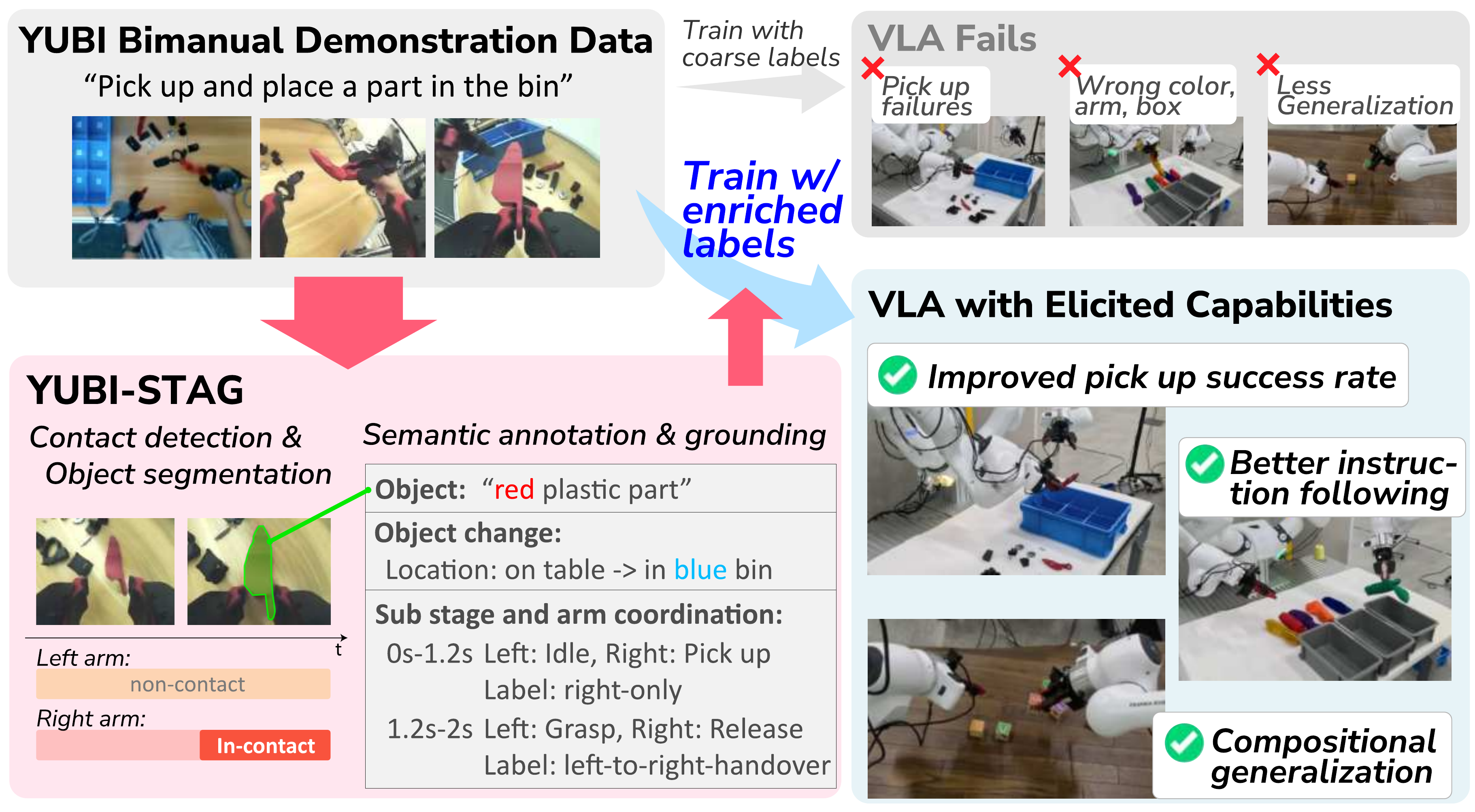}
  \caption{
  From a bimanual demonstration carrying only a coarse task label such as
  ``pick and place a part,'' YUBI-STAG recovers contact-anchored annotations of
  object identity, per-gripper action, bimanual coordination, and object change,
  which post-training uses to align the pretrained capabilities of a VLA with
  fine-grained manipulation language.}
  \label{fig:teaser}
\end{figure}

Existing VLA post-training pipelines rely on coarse episode-level language
instructions and unsegmented action sequences~\citep{o2024open,khazatsky2024droid}.
This coarse supervision fails to convey spatio-temporal dynamics, such as
gripper-object contact transitions, grip affordances, and bimanual coordination,
required for contact-rich, high-precision manipulation.
Inspired by the \emph{Superficial Alignment Hypothesis} in LLMs~\citep{zhou2023lima},
which posits that pre-trained models already possess extensive world knowledge
and post-training primarily elicits proper output formats, we argue that a
similar paradigm holds for VLAs.
Pre-trained VLAs already encode rich geometric and physical representations;
their limitation is not innate capability but the lack of precise alignment
signals during post-training~\citep{
chen2024sharegpt4v,yu2024rlaifv}.
\revised{Existing instruction augmentation and intermediate representations
either describe execution in language~\citep{xiao2023dial,blank2024nils,belkhale2024rth,hu2026finevla}
or ground actions and spatial plans in image coordinates~\citep{zawalski2024ecot,yuan2024robopoint,li2025hamster,lee2025molmoact,li2026robointer}.
None jointly recovers which gripper contacts which object, at which pixels, over
which interval, and how the two grippers coordinate from moving wrist views,
then uses this structure to align pretrained VLAs.}

We present \textbf{YUBI-STAG}, a framework that resolves this mismatch by
annotating demonstrations with the interaction structure they already contain,
and then post-training a VLA on those annotations.
Spatio-Temporal Annotation and Grounding (STAG) pairs a trained contact-object
segmentation module with a staged annotator built on an off-the-shelf multimodal
LLM~\citep{qwen36_27b}.
The segmentation module localizes contact intervals and contacted-object masks
from both wrist views.
The resulting \emph{contact tracks} identify when each gripper interacts with
which object, allowing YUBI-STAG to chain localized visual evidence across
interactions to recover object identities, per-gripper actions, bimanual
coordination, and state changes.
\revised{Because YUBI-STAG requires a temporally localized action sequence
and many model calls, we distill its structured annotation capabilities into
\textbf{YUBI-VLM}, aligning a pretrained VLM with the fine-grained interaction
semantics of YUBI data. YUBI-VLM directly annotates raw, unsegmented video
without predefined action boundaries and from wrist views alone, enabling
portable data collection. We evaluate its annotation capabilities on our new
benchmark, \textbf{YUBI-STAG-Bench}.}

Finally, we evaluate \emph{contact- and semantic-rich alignment} of VLAs using
the STAG annotations.
We pre-train a $\pi_{0.5}$-based policy~\citep{black2025pi05} on
\DBhours~hours of YUBI data\revised{, then post-train it with PA
instructions enriched by task-relevant STAG annotations, including object
identities, attributes, target locations, and spatial relations. We also
provide explicit handedness and contact-phase labels (approach, grasp, hold,
release) and use contact-prioritized sampling.}
Experiments on a bimanual Franka robot show that alignment with these rich
annotations improves fine-grained language following, contact-robust
manipulation, and compositional generalization to complex tasks.

The contributions of this paper are:
\begin{itemize}
  \item \textbf{YUBI-STAG and YUBI-VLM (\Cref{sec:stag}).} A contact-anchored annotation
  pipeline that turns coarsely labeled bimanual demonstrations into grounded
  spatio-temporal annotations, and a distilled model that produces them directly
  from raw, unsegmented, wrist-view-only video without action-boundary
  supervision.

  \item \textbf{YUBI-STAG-Bench (\Cref{sec:stag-bench}).} \revised{A benchmark of structured annotation quality on unseen bimanual tasks, covering contact intervals and object masks, PA segmentation and descriptions, object identities and attributes, coordination, spatial relations, and state transitions.}

  \item \textbf{Contact- and semantic-rich VLA post-training (\Cref{sec:vla-training}).} A contact and semantic-rich post-training recipe supplying handedness, contact phases, and
  contact-aware sampling to a $\pi_{0.5}$-based policy, extending data-centric
  alignment from instruction text to physical interaction. 
\end{itemize}

\section{Related Work}

\subsection{Scalable Collection of Unconstrained Manipulation Data}
Recent advances in low-cost bimanual systems~\citep{zhao2023learning,fu2024mobilealoha},
teleoperation~\citep{qin2023anyteleop}, and portable handheld
grippers~\citep{chi2024universal,ohkawa2026yubi} have enabled large-scale
collection of diverse robot demonstrations~\citep{o2024open,khazatsky2024droid,walke2023bridgedatav2}.
As collection scales toward in-the-wild settings, however, demonstrations
increasingly contain background clutter, camera motion, operator variation,
and lack explicit action boundaries~\citep{mandlekar2022what,bharadhwaj2024roboagent,cui2023play}.
Converting such noisy, unsegmented data into fine-grained supervision therefore
requires substantial curation and annotation, creating a new bottleneck for
scalable robot learning.

\subsection{Video-Language Grounding and Automated Data Annotation}
\revised{Foundation models have enabled scalable annotation of robot
data~\citep{chen2024sharegpt4v,qwen36_27b,lin2024videollava}.
One line relabels demonstrations with finer-grained language, including
execution details such as the active arm and contact
region~\citep{xiao2023dial,blank2024nils,belkhale2024rth,smith2024steer,hu2026finevla},
but does not jointly localize gripper--object contact in pixels and time.
Another grounds intermediate representations such as boxes, traces, and
contact points in images~\citep{zawalski2024ecot,li2025hamster,lee2025molmoact,li2026robointer}.
SPARC~\citep{blank2026sparc} and Robo2VLM~\citep{chen2025robo2vlm} derive grasp
or manipulation phases from gripper state and other proprioceptive cues, which
can miss non-grasping contacts.
In contrast, YUBI-STAG detects gripper--object contact visually, akin to
hand--object contact~\citep{shan2020understandinghands} and its
fine-grained dynamics~\citep{tateno2026handyvqa}, and recovers
per-gripper contact intervals, contacted-object masks, and bimanual
coordination from moving wrist views. YUBI-VLM uses the resulting contact
tracks to annotate raw, wrist-only video.}

\subsection{Data-Centric Post-Training Alignment in VLA Policies}
In NLP, the Superficial Alignment Hypothesis~\citep{zhou2023lima} posits that
vast knowledge and reasoning capability are acquired during pre-training, whereas post-training alignment primarily elicits output formats and behavioral styles using
high-quality supervision signals.
Its strong form is contested: stylistic alignment is recoverable from the base
model~\citep{lin2024unlocking}, while task competence continues to scale with
post-training data~\citep{raghavendra2024revisiting}.
While this data-centric alignment paradigm is well-known in LLM/VLMs, its
extension to Vision-Language-Action (VLA)
models~\citep{brohan2023rt2,kim2024openvla,octo2024,black2025pi05} has only
recently begun to be explored.
Existing VLA alignment efforts optimize preferences~\citep{zhang2024grape},
curate which demonstrations to train on~\citep{chen2025curating}, or pair fixed
trajectories with fine-grained instructions~\citep{hu2026finevla}.
However, these efforts, like standard behavior cloning, provide no explicit
supervision at contact transitions (approach, grasp onset, hold, release) or
bimanual coordination moments.

\section{Spatio-Temporal Interaction Annotation}
\label{sec:stag}

To promote alignment between VLA policies and fine-grained manipulation
language, we introduce two complementary annotation approaches. We propose
\textbf{YUBI-STAG} (\Cref{subsec:stag}), a Spatio-Temporal Annotation and Grounding (STAG) pipeline
that enriches YUBI demonstrations with contact-aware, rich annotations grounded in space and time. We distill this multi-stage pipeline into \textbf{YUBI-VLM} (\Cref{subsec:yubi-vlm}), which directly annotates raw videos at lower cost and supports wrist-only inference for portable data collection.

\subsection{YUBI-STAG}
\label{subsec:stag}
YUBI demonstrations provide temporally localized primitive-action (PA) labels
such as ``Open the lid of the jar.''  Following YUBI terminology, a PA denotes
a composite manipulation unit, and we call its constituent actions (e.g.,
reach, grasp, rotate, and release) \emph{PA stages}.  The original annotations
label only the coarse intent of the entire PA; they provide neither its stage
boundaries and actions nor object attributes, bimanual coordination, or
resulting object changes.  YUBI-STAG takes the PA labels and synchronized
videos from one overhead and two wrist cameras, and recovers the missing
elements as spatio-temporally grounded, structured annotations that can be
readily added as language supervision for VLA training.

\begin{figure*}[thpb]
   \centering
   \includegraphics[width=\textwidth]{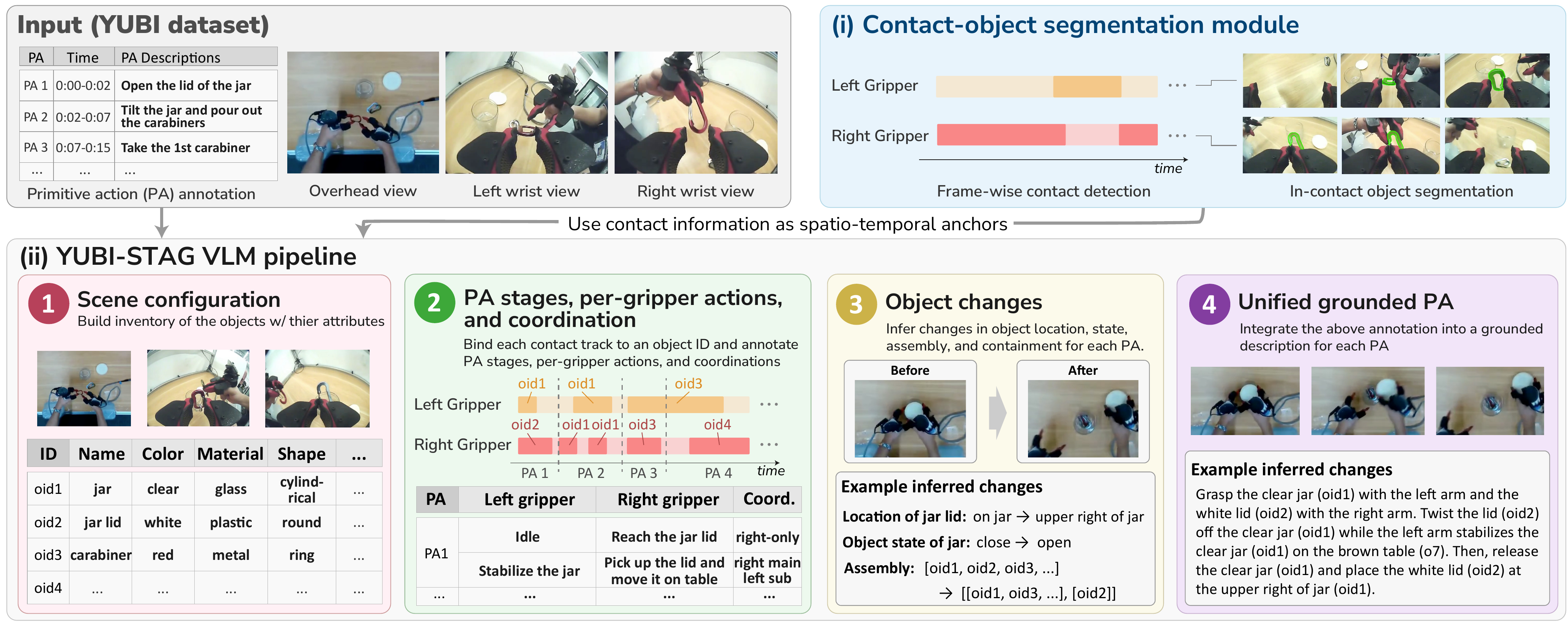}
   \caption{
   \textbf{Overview of YUBI-STAG.}
The contact-object segmentation module extracts \emph{contact tracks}---contact intervals with per-frame object masks---as spatio-temporal anchors for four annotation stages: scene configuration, PA stages and gripper coordination, object changes, and grounded PA descriptions.
   }
   \label{fig:stag_pipeline}
\end{figure*}

YUBI-STAG combines (i) a contact-object segmentation module and (ii) a staged
VLM-based annotator; \Cref{fig:stag_pipeline} provides an overview, and the four
annotation groups are detailed below.  The given PA intervals serve as temporal
scaffolds, while retaining PAs in the output schema allows YUBI-VLM to later
learn the complete schema from raw wrist-camera videos without predefined PAs.

\noindent
\textbf{(i) Contact-object segmentation module}
Fine-grained annotation requires knowing when each gripper contacts which
object.  A common proxy detects grasps from changes in gripper aperture, but it
is unstable and misses non-grasping contacts used to stabilize or support
objects.  Because low-cost systems such as UMI and YUBI lack tactile sensors,
we instead infer contact and the contacted object visually.

Our module encodes both wrist-camera streams with a frozen
DINOv2-L~\citep{oquab2023dinov2} and jointly models their frame features using a
lightweight temporal transformer.  For each gripper, an MLP head predicts the
contact state and a DPT-style decoder predicts the pixel-level mask of the
contacted object.

To train at scale, we derive pseudo-labels from gripper-aperture cues.  When the
aperture stabilizes, a point between the fingertips prompts SAM2~\citep{ravi2025sam}
to segment and bidirectionally track a candidate object.  We filter background
candidates using point-tracking co-motion~\citep{karaev2025cotracker3} and
monocular-depth agreement, then manually filter the results for initial training.
Finally, we self-train on more diverse data and fine-tune on a small set of
precise human annotations.

\noindent
\textbf{(ii) VLM-based annotation pipeline}

Rather than infer the full annotation in one MLLM call, YUBI-STAG organizes
MLLM queries into four stages, each building on previously resolved annotations
and relevant visual evidence. The preceding contact-object segmentation module
provides \emph{contact tracks}, each consisting of a contact interval and
per-frame object masks. These tracks anchor when and where interactions occur.

\paragraph{1) Scene configuration.}
The first stage builds a scene-level inventory of object names and counts.
The MLLM jointly examines the overall scene and object-centric observations localized by contact masks, providing both global scene context and clear views of objects during interaction.
For each instance, it records color, material, shape, size, and role (part,
container, surface, or fixture) under a persistent ID shared by all later
annotations.
Establishing this inventory first avoids assigning new identities to the same physical object at different interactions.

\paragraph{2) PA stages, per-gripper actions, and coordination.}
Before annotating gripper actions, we bind each contact track to an object ID
from the inventory. Appearance alone is ambiguous when instances look alike,
so we also exploit temporal continuity: tracks separated by a release and
immediate re-grasp, or connected by a handover, should refer to the same object.
Combining these constraints with appearance within the track masks provides
stable object identities across interactions.
The given PA intervals provide coarse temporal structure, which we refine into
constituent stages using the contact tracks. For each stage, we record the
participating object IDs, each gripper's action, and one of seven coordination
types: left- or right-only, either gripper as primary with the other supporting,
either handover direction, or bimanual.

\paragraph{3) Object changes.}
We determine how each interaction changes the manipulated objects and their
relations, using wrist views for close-up evidence and the overhead view for
broader spatial context. Comparing each contacted object before contact and
after release, we record its source and destination relative to other objects,
together with changes in state, assembly, or containment.
\revised{Following action-based object-state reasoning~\citep{tateno2025learning},
we select state observations using PA descriptions and grasp intervals. The
MLLM then uses the current PA description, the previously observed state, and
the corresponding visual evidence to determine whether the object's state
changed during that PA.}

\paragraph{4) Unified grounded PA annotation.}
Finally, the MLLM combines the resolved PA stages, per-gripper actions,
coordination, and object changes into a grounded PA description, with each
object mention linked to its persistent ID.

\noindent
\textbf{The cost of the YUBI-STAG pipeline}, in terms of processing time, the number of requests, and output token count, is discussed in the experiments section (\Cref{subsec:stag-bench-results}).

\subsection{Training YUBI-VLM}
\label{subsec:yubi-vlm}

We distill YUBI-STAG to reduce its multi-stage MLLM annotation cost and adapt a
pretrained VLM to YUBI videos and annotation conventions, particularly the
close-range views from moving wrist cameras.

\noindent\textbf{Architecture and annotation pipeline.}
YUBI-VLM is initialized from Qwen3.6-27B~\citep{qwen36_27b} and trained on ten
\revised{prediction modes covering scene inventory and attributes; PA span
localization from given descriptions or joint PA description and span
generation; held-object assignment; PA sub-stages, coordination, and
descriptions; PA-level object states; state-axis discovery and per-PA semantic
states; containment and attachment; and grounded PA descriptions.}
Contact-object masks are overlaid as contours on wrist frames, with gripper and
temporal information provided as text, enabling spatio-temporal grounding
without additional mask encoders or visual tokens.  Wrist frames are sampled
preferentially around contact intervals.

\noindent\textbf{Training.}
We jointly train all prediction modes on YUBI-STAG annotations using LoRA on
Qwen3.6-27B (489M trainable parameters; 1.76\%).  To support wrist-only
inference, the overhead view is dropped from 50\% of samples, while temporal
cropping and variable frame-rate sampling cover both short manipulation steps
and long videos.

\noindent\textbf{Training data.}
Training uses 2,477 sessions from 129 long-horizon tasks (29.4 hours; 18,336 PA
clips), yielding 29,500 examples.  All ten YUBI-STAG-Bench tasks are excluded
to evaluate generalization to unseen tasks.

\section{YUBI-STAG-Bench}
\label{sec:stag-bench}

To systematically evaluate structured annotations from our VLM-based pipeline,
we introduce \textbf{YUBI-STAG-Bench}, a benchmark for semantically rich,
spatio-temporally grounded understanding of long-horizon robot manipulation.
It contains 100 episodes from 10 tasks (10 per task), manually annotated
following the YUBI-STAG schema.

We densely annotate fields central to downstream robot learning and amenable to
structured evaluation: \emph{objects}, \emph{attributes}, \emph{contact tracks},
\emph{primitive actions (PAs)}, \emph{coordination types}, \emph{object
locations}, and \emph{object states}.
The benchmark contains \textbf{510 objects}, \textbf{970 PAs}, and
\textbf{2,201 interaction intervals}, with per-gripper temporal intervals and
contacted-object masks annotated at 10 fps.

We organize the benchmark into two task groups.  \textbf{Contact-object
grounding} detects each gripper’s contact intervals from wrist video and
segments the contacted object at 10 fps.  \textbf{Semantic annotation}
evaluates structured manipulation understanding from video, task
description, and contact tracks.  \emph{Object ID} identifies manipulated
objects, and \emph{Attribute} predicts shape and material.  \emph{TAS}
localizes a given sequence of PAs, while \emph{PA GEN TAS} jointly predicts PA
descriptions and intervals from unsegmented video.  \emph{Coordination}
predicts bimanual roles, \emph{Location} predicts spatial relations and
reference objects, and \emph{State} predicts object states and their transition
times.  Together, these tasks evaluate the semantic and spatio-temporal
annotations produced by YUBI-STAG or YUBI-VLM.

\section{Training VLA Policies}
\label{sec:vla-training}
\begin{figure*}[t]
   \centering
   \includegraphics[width=\textwidth]{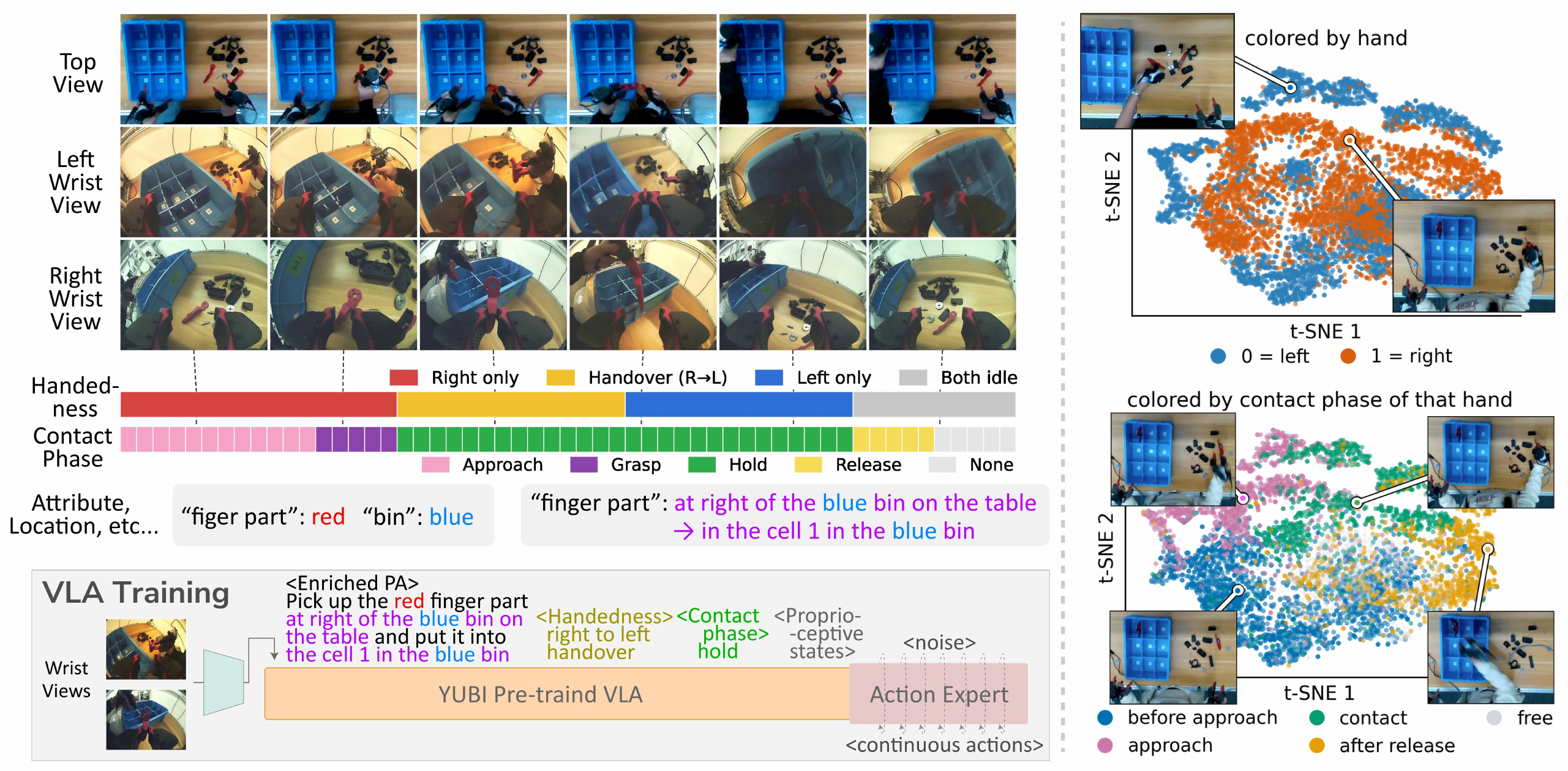}
   \caption{\textbf{Contact- and semantic-rich post-training for YUBI-VLA.}
   \emph{Left:} YUBI-STAG augments primitive-action instructions with
   \revised{task-relevant object attributes and locations, together with
   time-varying handedness} and contact-phase labels while the unchanged
   flow-matching expert learns continuous bimanual actions.
   \emph{Right:} t-SNE projections of per-hand delta actions after PCA-50
   reduction, colored by hand identity and contact phase, reveal
   \revised{handedness and contact-phase} structure in the trajectories and
   motivate the proposed labels.}
   \label{fig:yubi_vla}
\end{figure*}

We evaluate whether YUBI-STAG annotations improve VLA training while isolating
the effect of interaction-rich supervision from policy design.
We adopt a $\pi_{0.5}$-based VLA~\citep{black2025pi05}, retaining its
vision-language backbone and flow-matching action expert.
The policy is first pretrained on the large-scale YUBI corpus
(\DBhours hours~\citep{ohkawa2026yubi}) and then post-trained on the same small
set of demonstrations (10--20~hours) under different STAG-derived supervision.
Thus, our ablations differ only in the supervision used to elicit the shared
pretrained manipulation capability.  The pipeline is shown in~\Cref{fig:yubi_vla}.
\subsection{YUBI VLA and Action Representation}
At time $t$, the policy takes two wrist images, language, and proprioception
and predicts a horizon of bimanual actions $\mathbf{a}_{t:t+H-1}$.
Each action contains a 6-DoF delta end-effector motion and gripper command for
each YUBI gripper.  We use the UMI end-effector state
representation~\citep{ohkawa2026yubi,chi2024universal}, making demonstrations
collected with the hand-held interface directly usable for policy training.
Following $\pi_{0.5}$, the action expert learns an action chunk by conditional
flow matching~\citep{lipman2023flow,liu2023rectified}.  Given context $c_t$, it
transports Gaussian noise $\symbf{\epsilon}$ toward the demonstrated
action $\mathbf{a}$ using
\begin{equation}
\mathcal{L}_{\mathrm{FM}} =
\mathbb{E}_{\mathbf{a},\symbf{\epsilon},s}
\left[\left\|v_\theta(\mathbf{a}_s,s,c_t)
- (\mathbf{a}-\symbf{\epsilon})\right\|_2^2\right],
\end{equation}
where $\mathbf{a}_s=(1-s)\symbf{\epsilon}+s\mathbf{a}$ and $s\in[0,1]$
is the flow time.  Apart from the supervision described below, we follow the
standard $\pi_{0.5}$ recipe without modifying the network or objective.
\subsection{Contact- and Semantic-Rich Post-Training}
Original YUBI annotations provide episode-level goals and primitive actions
(PAs), but not how each interaction is executed.
\revised{As illustrated in~\Cref{fig:yubi_vla}, we enrich PA instructions
with task-relevant object and spatial information from YUBI-STAG. We also
propose handedness and contact-phase labels as default, broadly applicable
additions.}
\revised{These annotations provide more precise language supervision without
changing the observation or action space.}
\paragraph{\revised{Task-specific PA enrichment.}}
\revised{Depending on the task, we enrich PA descriptions with
STAG-derived object identities and attributes, target locations, and spatial
relations. For example, part sorting specifies the part's color and destination
bin cell.}
\paragraph{\revised{Handedness labels.}}
\revised{We augment each PA with the acting gripper or handover direction}, e.g.,
\texttt{\{PA\} with the left gripper}.  The four types are left gripper,
right gripper, left-to-right handover, and right-to-left handover.
Unlike coarse PA or episode instructions, these labels identify the responsible
gripper and are updated with STAG stages over time.
\paragraph{Contact-phase labels.}
We further label the relevant gripper--object interaction based on STAG's contact segments as
\emph{approach}, \emph{grasp}, \emph{hold}, or \emph{release}, e.g.,
\texttt{\{PA\}, approaching it}.  
\emph{Grasp} spans $\pm$0.2\,s around contact onset, \emph{release} the last 0.2\,s of the contact segment, and the remaining contact frames are \emph{hold}.
These labels ground language in the current phase of the physical interaction rather than only its goal.
\revised{Handedness and contact-phase labels} can be used independently or jointly, as shown
in~\Cref{fig:yubi_vla}.
\paragraph{Contact-prioritized sampling.}
Because contact transitions are sparse but often critical to task success, we
rebalance training chunks toward contact phases, as the principle of
prioritized replay~\citep{schaul2016prioritized}.  For example, a grasp-onset
phase comprising 5\% of demonstrations can constitute 30\% of sampled chunks.
This exposes approach, grasp, hold, and release transitions at useful frequency
without altering the training objective.
The same dense annotations also support data curation by identifying
instruction--trajectory mismatches and diversifying language descriptions.
Overall, our data-centric post-training keeps the observations, actions,
architecture, and objective fixed, isolating the effect of supervision that
captures both interaction semantics and temporal contact structure.

\section{Experiments}
\label{sec:experiments}

\revised{We first evaluate the annotation quality and inference efficiency
of YUBI-STAG and YUBI-VLM on YUBI-STAG-Bench
(\Cref{subsec:stag-bench-results}). We then assess the effects of handedness
and contact supervision on fine-grained manipulation
(\Cref{subsec:fine-manip-results}) and test language control over object
identity, acting gripper, and spatial targets using enriched PAs
(\Cref{subsec:language-steerability}). Finally, we evaluate action composition
from YUBI-VLM-generated PAs on seen and unseen tasks
(\Cref{subsec:action-composition}).}

\subsection{Results on YUBI-STAG-Bench}
\label{subsec:stag-bench-results}

\noindent\textbf{Evaluation Protocol.}
Contact detection uses temporal mIoU and frame accuracy per episode--gripper
pair; contact masks use DAVIS $J$ and $F$~\revised{\citep{perazzi2016davis}}
per ground-truth track, with uncovered
frames scored as zero.  Semantic objects are matched one-to-one by contact-track
overlap, without using predicted names.  As in standard action segmentation,
TAS uses boundary F1@0.5\,s and segment IoU;
\revised{PA GEN TAS additionally measures PA-count accuracy and evaluates
generated descriptions using GPT-5.2 on PAs matched one-to-one by temporal IoU.
The judge assesses whether the description refers to the same step despite
wording differences (identity), specifies the object, action, and any required
destination to reproduce it from text alone (sufficiency), and covers exactly
one reference PA rather than merged or partial steps (granularity). We report
the percentage of matched pairs satisfying each criterion, with identity and
sufficiency judged independently.}
We report accuracy for object identity, attributes,
coordination, and locations, and temporal mIoU and transition F1@1.0\,s for
states; all scores are percentages.

\noindent\textbf{Contact-Object Grounding.}
We evaluate the contact-object segmentation module on grasp intervals from 100 benchmark episodes across both grippers.
The baseline uses absolute-aperture cues to detect grasp intervals from gripper closing and subsequent aperture stabilization, then applies SAM2 within these intervals.

\Cref{tab:contact-results} shows that our module improves both temporal detection and spatial grounding across all ten tasks.
These gains support learned contact tracks as reliable spatio-temporal anchors for YUBI-STAG.

\begin{table}[t]
  \centering
  \small
  \setlength{\tabcolsep}{5pt}
  \setlength{\aboverulesep}{2pt}
  \setlength{\belowrulesep}{2pt}
  \caption{
    Grasp-based contact-object grounding on YUBI-STAG-Bench.
    F-Acc.: frame accuracy; e2e: all ground-truth mask frames,
    with uncovered frames scored as zero.
  }
  \label{tab:contact-results}
  \begin{tabular}{@{}l|cc|cc@{}}
    \toprule
    &
    \multicolumn{2}{c|}{\textbf{Grasp Detection}} &
    \multicolumn{2}{c}{\textbf{Contact-Obj. Seg.}} \\
    \cmidrule(lr){2-3}
    \cmidrule(lr){4-5}
    \textbf{Method} &
    \textbf{mIoU $\uparrow$} &
    \textbf{F-Acc. $\uparrow$} &
    \textbf{$J$ (e2e) $\uparrow$} &
    \textbf{$F$ (e2e) $\uparrow$} \\
    \midrule[\heavyrulewidth]

    Aperture + SAM2
      & 71.8 & 83.8 & 40.3 & 43.6 \\

    Ours
      & \textbf{89.7} & \textbf{94.5} & \textbf{74.0} & \textbf{75.1} \\

    \bottomrule
  \end{tabular}
\end{table}

\noindent\textbf{Semantic Annotation.}
We evaluate semantic annotation while varying PA input and camera configuration.
\textbf{+PA} provides ground-truth PA descriptions and intervals, \textbf{+PA
list only} provides descriptions but requires interval prediction, and the
default setting uses raw, unsegmented video; \textbf{w/o Top} retains only the
wrist views. We apply the same pipeline and parser to the untrained Qwen3.6-27B
backbone to isolate the effect of distillation; it also receives a
task-independent formatting example, and all methods complete all 100 episodes.
We also evaluate RoboFine-VLM-397B-A17B, the annotator released with
FineVLA~\citep{hu2026finevla}, which fine-tunes a 397B-parameter
mixture-of-experts VLM on robot manipulation videos, by replacing the
Qwen3.6-27B backbone under the same evaluation setup.

\begin{table*}[t]
  \centering
  \scriptsize
  \setlength{\tabcolsep}{2.0pt}
  \setlength{\aboverulesep}{1.5pt}
  \setlength{\belowrulesep}{1.5pt}
  \caption{
    Semantic annotation results on YUBI-STAG-Bench.
    \textbf{B-F1}: boundary F1 at 0.5\,s;
    \textbf{S-IoU}: segment IoU;
    \textbf{\#Acc.}: PA-count accuracy;
    \textbf{Id.}/\textbf{Suf.}/\textbf{Gr.}: semantic identity, sufficiency, and granularity;
    \textbf{5-/7-way}: coordination accuracy;
    \textbf{Cont.}/\textbf{Rel.}: container and relative-location accuracy;
    \textbf{T-F1}: state-transition F1 at 1.0\,s.
    \textbf{+PA} uses ground-truth PA annotations as input;
    \textbf{+PA list only} uses only the ground-truth PA descriptions and
    predicts their temporal intervals;
    \textbf{w/o Top} uses only the two wrist views.
    Evaluation scores are reported in \%; \textbf{Speed} is video duration
    divided by wall-clock time on one eight-H200 node.
    \textbf{Bold}: best; \underline{underline}: second best.
  }
  \label{tab:stag-bench-results}

  \resizebox{\textwidth}{!}{
  \begin{tabular}{@{}l|c|c|cc|cc|cccccc|cc|cc|cc@{}}
    \toprule
    &
    &
    \textbf{Object ID} &
    \multicolumn{2}{c|}{\textbf{Attribute}} &
    \multicolumn{2}{c|}{\textbf{TAS}} &
    \multicolumn{6}{c|}{\textbf{PA GEN TAS}} &
    \multicolumn{2}{c|}{\textbf{Coordination}} &
    \multicolumn{2}{c|}{\textbf{Location}} &
    \multicolumn{2}{c}{\textbf{State}} \\
    \cmidrule(lr){3-3}
    \cmidrule(lr){4-5}
    \cmidrule(lr){6-7}
    \cmidrule(lr){8-13}
    \cmidrule(lr){14-15}
    \cmidrule(lr){16-17}
    \cmidrule(lr){18-19}

    \textbf{Method} &
    \textbf{Speed} &
    \textbf{Acc.} &
    \textbf{Shape} & \textbf{Material} &
    \textbf{B-F1} & \textbf{S-IoU} &
    \textbf{B-F1} & \textbf{S-IoU} & \textbf{\#Acc.} &
    \textbf{Id.} & \textbf{Suf.} & \textbf{Gr.} &
    \textbf{5-way} & \textbf{7-way} &
    \textbf{Cont.} & \textbf{Rel.} &
    \textbf{mIoU} & \textbf{T-F1} \\
    \midrule[\heavyrulewidth]

    \multicolumn{19}{@{}l}{\grouplabel{Baseline: untrained backbone}} \\
    Qwen3.6-27B + PA
      & 3.9$\times$ & \underline{96.6} & 64.0 & 85.5 & -- & -- & -- & -- & -- & -- & -- & -- & 76.8 & 51.0 & 54.4 & 31.4 & 43.3 & 11.9 \\
    Qwen3.6-27B + PA list only
      & 3.9$\times$ & 94.2 & 67.1 & 83.7 & 24.0 & 55.2 & -- & -- & -- & -- & -- & -- & 73.8 & 47.2 & 51.0 & 25.0 & 41.6 & 15.9 \\
    Qwen3.6-27B
      & 3.1$\times$ & 91.5 & 58.0 & \underline{89.6} & -- & -- & 14.9 & 41.3 & 3.0 & 23.3 & 58.8 & 55.1 & 62.3 & 44.3 & 43.5 & 22.9 & 37.9 & 11.9 \\
    Qwen3.6-27B w/o Top
      & 3.5$\times$ & 92.0 & 58.7 & 88.3 & -- & -- & 16.1 & 45.2 & 10.0 & 29.5 & 66.3 & 64.2 & 68.8 & 46.7 & 16.9 & 14.9 & 38.4 & 12.3 \\
    \midrule
    \multicolumn{19}{@{}l}{\grouplabel{Baseline: VLM fine-tuned on robot manipulation videos}} \\
    RoboFine-VLM-397B + PA
      & 1.6$\times$ & 89.7 & 56.0 & 87.0 & -- & -- & -- & -- & -- & -- & -- & -- & 83.9 & \underline{55.7} & 66.7 & 47.9 & 50.9 & 15.9 \\
    RoboFine-VLM-397B + PA list only
      & 1.6$\times$ & 88.7 & 56.4 & 87.4 & \underline{36.8} & \underline{59.1} & -- & -- & -- & -- & -- & -- & 77.6 & 49.0 & 64.4 & 41.3 & 44.7 & 16.2 \\
    RoboFine-VLM-397B
      & 0.9$\times$ & 85.0 & 60.8 & 87.2 & -- & -- & 15.6 & 31.8 & 2.0 & 25.9 & 49.2 & 53.4 & 68.4 & 50.6 & 51.3 & 33.5 & 43.0 & 15.1 \\
    RoboFine-VLM-397B w/o Top
      & 1.2$\times$ & 89.2 & 60.2 & 84.4 & -- & -- & 15.0 & 30.7 & 0.0 & 26.9 & 49.1 & 51.2 & 62.6 & 46.3 & 44.3 & 27.0 & 44.2 & 14.0 \\
    \midrule
    \multicolumn{19}{@{}l}{\grouplabel{Ours: distilled VLM annotator}} \\
    YUBI-VLM + PA
      & 3.7$\times$ & \textbf{98.0} & 67.0 & 85.5 & -- & -- & -- & -- & -- & -- & -- & -- & \textbf{93.3} & 55.5 & \underline{96.0} & \textbf{64.0} & \underline{64.3} & \underline{64.5} \\
    YUBI-VLM + PA list only
      & 3.5$\times$ & \textbf{98.0} & \underline{67.7} & 85.6 & \textbf{82.7} & \textbf{83.9} & -- & -- & -- & -- & -- & -- & 91.7 & 54.1 & \textbf{96.7} & 58.9 & 62.2 & 58.4 \\
    YUBI-VLM
      & 3.5$\times$ & 95.2 & 66.4 & 83.7 & -- & -- & \underline{77.9} & \underline{78.4} & \underline{46.0} & \textbf{64.2} & \textbf{96.1} & \textbf{87.1} & 89.1 & 51.2 & \textbf{96.7} & \underline{63.9} & 52.0 & 43.5 \\
    YUBI-VLM w/o Top
      & 3.9$\times$ & 93.9 & 64.6 & 87.7 & -- & -- & \textbf{78.2} & \textbf{78.5} & \textbf{49.0} & \underline{59.9} & \underline{95.7} & \underline{86.0} & 90.6 & 52.7 & 90.0 & 50.0 & 47.9 & 31.0 \\
    \dashedrow{19}
    \multicolumn{19}{@{}l}{\grouplabel{Ours: multi-stage annotation pipeline}} \\
    YUBI-STAG \revised{(given PAs)}
      & 1.5$\times$ & 94.2 & \textbf{72.1} & \textbf{92.1} & -- & -- & -- & -- & -- & -- & -- & -- & \underline{93.0} & \textbf{61.6} & 95.3 & 55.9 & \textbf{76.0} & \textbf{67.9} \\
    \bottomrule
  \end{tabular}
  }
\end{table*}

As shown in~\Cref{tab:stag-bench-results}, YUBI-STAG performs strongly overall, with remaining challenges in fine-grained appearance, relative location, and state changes, which require visual details often missing from the distant top or moving wrist views. With matched PA input, YUBI-VLM preserves most of these capabilities and matches or exceeds YUBI-STAG on several categories, showing that distillation effectively aligns the pretrained VLM with YUBI videos and annotation conventions, although YUBI-STAG remains stronger on attributes and states.

Compared with the untrained backbone, YUBI-VLM substantially improves temporal, coordination, spatial, and state understanding while recovering PAs directly from raw video. It retains most performance with wrist views alone, with drops in relative location and state timing indicating the value of the top view for global context.
RoboFine-VLM, despite large-scale training on robot videos, remains well below YUBI-VLM. While capable of detailed action descriptions, it is not trained for temporal localization and tends to over-segment PAs, limiting its spatio-temporal grounding.

\noindent\textbf{Inference Cost.}
Under the same eight-H200 setup, YUBI-VLM processes all 100 episodes
(70.9 minutes of video) in less than half the wall-clock time of YUBI-STAG,
reaching 3.5–3.7$\times$ real-time throughput. Per episode, YUBI-STAG
requires 154 VLM requests, versus 33.0–34.1 for YUBI-VLM, a 4.5-fold
difference. YUBI-STAG generates 14.6k output tokens per
episode, whereas YUBI-VLM generates only 3.3–3.4k, reflecting its more
compact annotation pipeline. RoboFine-VLM is substantially
slower due to its much larger model size, making YUBI-VLM both the most
efficient and the most accurate annotator evaluated.

\subsection{Results on Fine-grained Manipulation}
\label{subsec:fine-manip-results}
\noindent\textbf{Setup.}
We evaluate \revised{handedness and contact supervision} on two fine-grained
manipulation tasks. In \emph{YUBI part sorting}, the robot picks a specified
YUBI finger part and places it in a designated bin cell. In
\emph{fuel-bracket bin picking}, it picks one of five industrial parts from a source box and transfers it to a target box, with a timeout of 2.5 minutes.
\revised{The two setups are shown in the upper two panels of~\Cref{fig:socks_exp}.
Both tasks involve small, irregularly shaped parts, making reliable grasping
itself challenging.}
Both use a Franka robot with the YUBI bimanual end-effector, operating at
10~Hz with eight-step action chunks.
We compare two coarse-label baselines: the \emph{$\pi_{0.5}$ baseline},
post-trained on YUBI from $\pi_{0.5}$, and \emph{YUBI-VLA},
pretrained on the full YUBI corpus and post-trained with task-specific coarse
instructions. We then augment YUBI-VLA with \revised{handedness labels
(\emph{+Handedness})} and additional contact-aware supervision
\revised{(\emph{+Handedness + Contact})}.

\begin{figure}[t]
    \centering
    \includegraphics[width=0.7\columnwidth]{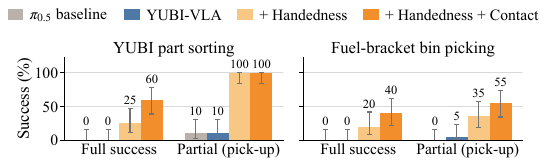}
    \caption{
    Fine-grained manipulation results over 20 trials per condition.
    Bars show success rates with 95\% Wilson score confidence intervals.
    Full success completes both picking and placement; partial success
    indicates successful pick-up.
    }
    \label{fig:fine-manip-results}
\end{figure}

\noindent\textbf{Results.}
\Cref{fig:fine-manip-results} shows consistent gains from richer supervision.
Neither coarse-label baseline completes either task. 
On YUBI part sorting,
\revised{handedness labels} yield 25\% full success and 100\% pick-up success, while adding
contact supervision further raises full success to 60\%.
This suggests that specifying both the acting gripper and interaction phase
improves precise manipulation beyond \revised{handedness supervision} alone.
Fuel-bracket bin picking is more challenging, requiring reliable acquisition
before placement.  \revised{Handedness supervision} achieves 20\% full success, which
contact-aware post-training doubles to 40\%, while also improving pick-up
success from 35\% to 55\%.
Across both tasks, progressively richer supervision improves the same policy
architecture, supporting our claim that contact-grounded, fine-grained
annotations better elicit manipulation skills acquired during large-scale YUBI
pretraining.

\revised{We observe more variable and hesitant motions in the coarse-label
baselines, whereas fine-grained supervision yields sharper motions. Projections
of demonstration delta actions in~\Cref{fig:yubi_vla} show distinct
regions associated with handedness and contact phase. These observations
suggest that aligning instructions with these action modes reduces ambiguity
in policy conditioning, helping the flow-matching expert learn more consistent
action flows.}

\subsection{Fine-grained Language Steerability}
\label{subsec:language-steerability}

\begin{figure}[t]
   \centering
   \includegraphics[width=0.9\columnwidth]{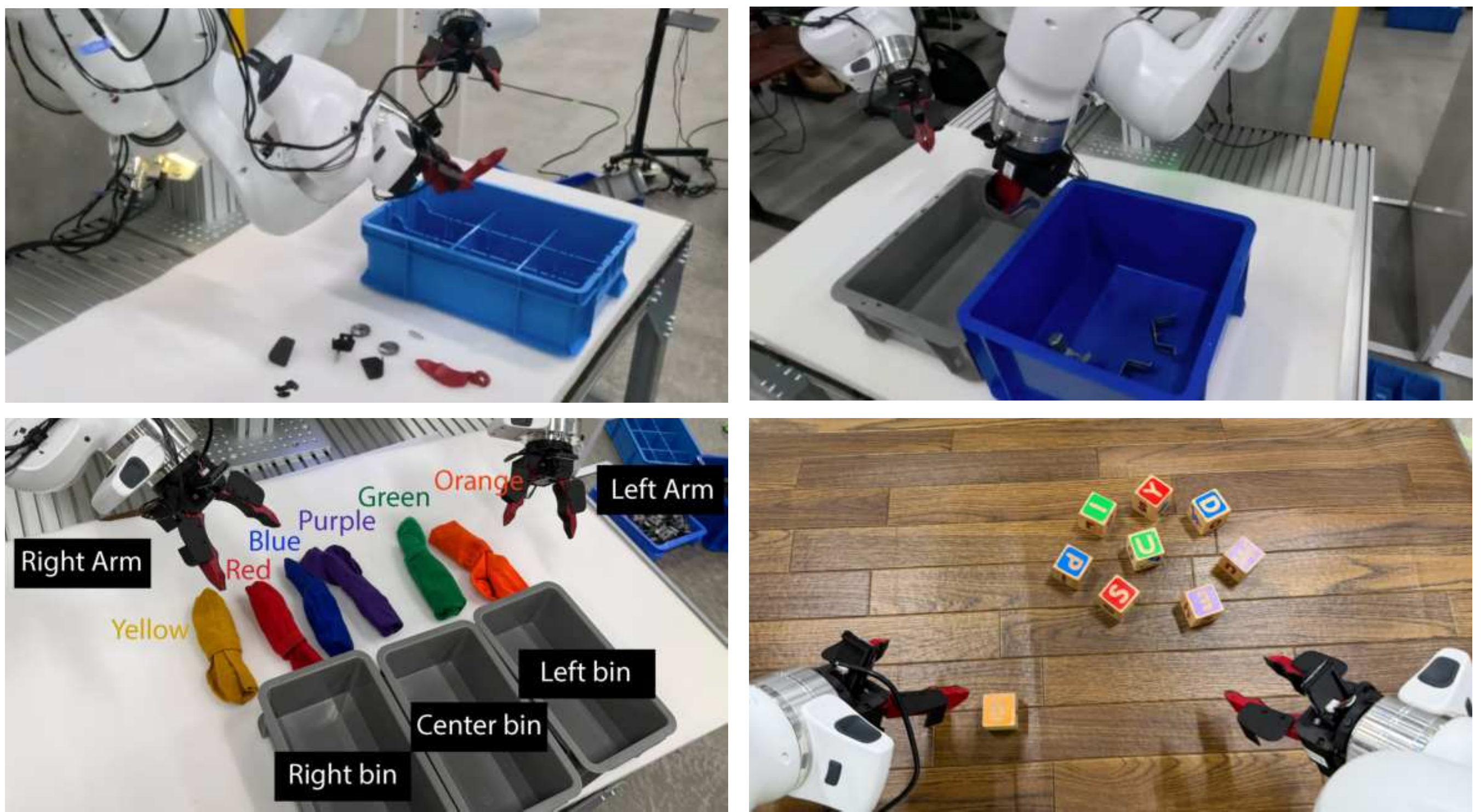}
   \caption{Experimental setup for robot evaluation tasks.
   }
   \label{fig:socks_exp}
\end{figure}

\noindent\textbf{Setup.}
We evaluate whether YUBI-STAG annotations enable fine-grained language
steerability beyond the semantics available in the original primitive
actions (PAs).
We consider a tabletop task with six differently colored socks and three
bins arranged from left to right, as shown in~\Cref{fig:socks_exp}.
In each demonstration, three randomly selected socks are placed into the
bins from left to right and subsequently removed and returned to the table.
The original PAs contain only coarse descriptions such as
``randomly pick up socks and put it into the box,'' without specifying
the sock identity, manipulation gripper, or fine-grained spatial target.
The selected sock colors are balanced across target bins, and the training
episodes are nearly balanced between the left (5,351; 48.3\%) and right
(5,723; 51.7\%) grippers.
YUBI-STAG enriches these coarse PAs with fine-grained semantics inferred
from the videos, which we use to train the manipulation policy.

\begin{figure}[t]
    \centering
    \begin{minipage}[t]{0.485\textwidth}
    \includegraphics[width=\linewidth]{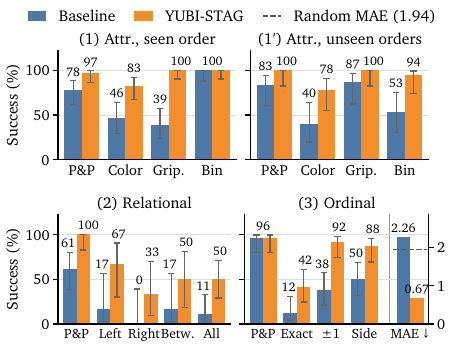}
    \caption{
Fine-grained language steerability.
Baseline (blue) and YUBI-STAG (orange) success rates with 95\% Wilson
confidence intervals.
P\&P in (3) does not measure instruction following because any sock may be grasped.
MAE: ordinal position error (right axis, lower is better), compared with
uniform random selection (dashed line).
Attr.: attribute; Grip.: gripper; Betw.: between.
}
    \label{fig:language-steerability}
    \end{minipage}\hfill
    \begin{minipage}[t]{0.485\textwidth}
    \includegraphics[width=\linewidth]{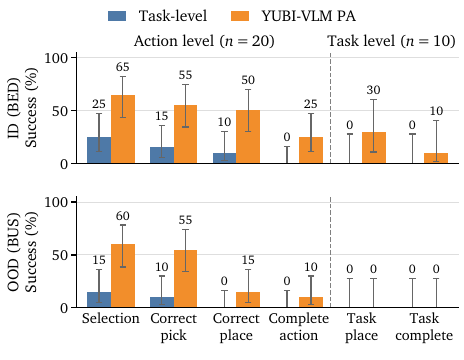}
    \caption{
    Letter-block action composition on BED (ID) and BUS (OOD).
    Bars compare the task-level baseline (blue) and YUBI-VLM PA supervision
    (orange) with 95\% Wilson score confidence intervals.
    Action metrics ($n=20$) are cumulative; task place and task complete
    require both actions to succeed through placement and orientation,
    respectively ($n=10$).
    }
    \label{fig:action-composition}
    \end{minipage}
\end{figure}

We evaluate three forms of language control:
(1) sock color, manipulation gripper, and target bin;
(2) relative placement with respect to other socks; and
(3) ordinal object selection, e.g., ``second from the left.''
In every experiment, evaluation trials are evenly divided across sock colors,
instructed grippers, and target bins.
We compare against a baseline policy trained without the enriched
YUBI-STAG annotations.
\revised{Both policies are post-trained from the pretrained YUBI-VLA using
the same training configuration as in~\Cref{subsec:fine-manip-results}, and
operate at 10~Hz with eight-step action chunks.}
The results are summarized in~\Cref{fig:language-steerability}.

\noindent\textbf{Color, gripper, and target-bin control.}
We first test joint control of sock color, manipulation gripper, and target bin.
In (1), both policies select the correct bin in all completed trials, while
YUBI-STAG achieves higher operation completion and color accuracy
(82.9\% vs.\ 46.4\%).
Its color errors mainly confuse red/\allowbreak orange/\allowbreak yellow and blue/\allowbreak purple, mirroring
errors in the generated annotations.
YUBI-STAG also follows the specified gripper in all completed trials, whereas
the baseline consistently uses the right gripper for left-gripper instructions
despite balanced post-training data.
This suggests that YUBI-STAG enables language-conditioned manipulation mode,
rather than retaining the pretraining right-gripper bias.

In (1'), we use two unseen bin orders, center--right--left and
right--left--center, requiring instruction following beyond the training
sequence.
YUBI-STAG largely retains its behavior, whereas baseline target-bin accuracy
drops from 100\% to 53.3\%, with its errors biased toward the left bin.
This suggests that the baseline relies on the visual scene and learned sequence
rather than grounding the target-bin instruction in language.

\noindent\textbf{Relational spatial placement.}
We next evaluate \textit{left of}, \textit{right of}, and \textit{between}
relations to color-specified reference socks.
YUBI-STAG achieves 50.0\% correct placement versus 11.1\% for the baseline.
Remaining errors are near-misses involving reference-color confusion,
left/right reversal, or placement on the reference.
\revised{Despite these annotation imperfections, the gain over the baseline shows that
the relational annotations are effectively reflected in policy behavior.}

\noindent\textbf{Ordinal spatial selection.}
Finally, we remove color cues and specify targets solely by ordinal position,
e.g., ``second from the left.''
\revised{In panel (3) of~\Cref{fig:language-steerability}, P\&P measures
pick-and-place completion with any sock. Exact requires the instructed sock,
$\pm$1 allows an ordinal error of at most one position, and Side checks whether
the selected sock is in the correct left or right half of the six-sock row.
These four rates use 24 trials each. The MAE bars use the right axis and report
the mean absolute difference between selected and target ranks over trials
with a recorded selection; lower is better. The dashed line marks the expected
error of uniform random selection (1.94 positions).}
YUBI-STAG achieves 41.7\% exact selection versus 12.5\% for the baseline;
91.7\% of its selections are nevertheless within one position of the target.
\revised{Its MAE is 0.67 positions, compared with 2.26 for the baseline.}
In contrast, the baseline frequently selects the leftmost sock and performs
worse than random selection in position error, indicating a strong positional
bias rather than ordinal instruction following.
Its comparable P\&P rate does not indicate \revised{correct ordinal selection}, as it can
simply grasp an easier sock.

Overall, YUBI-STAG enables language control over semantic, embodiment, and
spatial concepts absent from the original coarse PAs.
Remaining errors are structured: color errors reflect annotation confusions,
while spatial errors generally remain close to the intended target.

\subsection{Action Composition from YUBI-VLM-Generated PAs}
\label{subsec:action-composition}

\noindent\textbf{Setup.}
We test whether YUBI-VLM-generated PAs enable action composition for unseen
long-horizon tasks.
Using demonstrations of six letter-block arrangements
(\emph{BED}, \emph{LEFT}, \emph{NEW}, \emph{PUSH}, \emph{TUBE}, and
\emph{YUBI}), we compare a \emph{task-level} baseline conditioned on the
original long-horizon task instruction with \emph{YUBI-VLM PA supervision}, which segments
each demonstration using PA GEN TAS and conditions each segment on its generated
PA description.
\revised{Both policies are post-trained from the pretrained YUBI-VLA using
the same training configuration as in~\Cref{subsec:fine-manip-results}, and
operate at 10~Hz with eight-step action chunks.}

At evaluation, with \emph{B} pre-placed, the robot places the remaining two
letters for \emph{BED} (ID) and the unseen \emph{BUS} (OOD), with 10 rollouts
per method and task.
We cumulatively evaluate whether the robot selects the instructed letter
(target selection), lifts it (pick-up), places it at the correct location
(placement), and leaves it face-up with the letter upright (orientation;
rolled-over or sideways letters fail);
task-level success requires both actions to reach the criterion,
with failures and unreached stages counted as unsuccessful.

\noindent\textbf{Results.}
As shown in~\Cref{fig:action-composition}, YUBI-VLM PA supervision consistently
improves the action-success funnel on BED, particularly target selection and
placement, and enables task-level placement and completion never achieved by
the task-level baseline.
This demonstrates that action-specific language provides more effective
supervision than repeatedly conditioning on the coarse task instruction.
These gains transfer to the unseen BUS composition, with substantial
improvements in target selection and pick-up and successful placement and
completion of individual actions, which the baseline never achieves.
This suggests that generated PAs capture reusable action semantics that can be
recombined beyond training tasks.

\section{Conclusions}

We presented YUBI-STAG for contact-anchored, spatio-temporal annotation of bimanual demonstrations, and YUBI-VLM for directly annotating raw and wrist-only videos, with its annotation capability validated by YUBI-STAG-Bench on unseen manipulations.
Post-training YUBI-pretrained VLAs with these annotations
\revised{aligns them with fine-grained language, better eliciting their
pretrained capabilities and improving} fine-grained manipulation, language
following, and compositional manipulation beyond the original coarse labels.
\revised{YUBI-STAG and YUBI-VLM bridge coarse robot demonstrations and
fine-grained language alignment, enabling more capable and controllable VLAs.}

\section*{Acknowledgments}
This paper is based on results obtained from a project, JPNP25015, commissioned by the New Energy and Industrial Technology Development Organization (NEDO).

\bibliographystyle{plainnat}
\bibliography{IEEEabrv,venues_short,refs}

\end{document}